# A transparency-based action model implemented in a robotic physical trainer for improved HRI

Naama Aharony (N.A.)*

Department of Industrial Engineering & Management, Ben-Gurion University of the Negev, Beer-Sheva, Israel. naamaah@post.bgu.ac.il

Maya Krakovski (M.K.)*

Department of Industrial Engineering & Management, Ben-Gurion University of the Negev, Beer-Sheva, Israel. mayakrak@post.bgu.ac.il

Yael Edan (Y.E.)

Department of Industrial Engineering & Management, Ben-Gurion University of the Negev, Beer-Sheva, Israel. yael@bgu.ac.il

Transparency is an important aspect of human-robot interaction (HRI), as it can improve system trust and usability leading to improved communication and performance. However, most transparency models focus only on the amount of information given to users. In this paper, we propose a bidirectional transparency model, termed a transparency-based action (TBA) model, which in addition to providing transparency information (robot-to-human), allows the robot to take actions based on transparency information received from the human (robot-of-human and human-to-robot). We implemented a three-level (High, Medium and Low) TBA model on a robotic system trainer in two pilot studies (with students as participants) to examine its impact on acceptance and HRI. Based on the pilot studies results, the Medium TBA level was not included in the main experiment, which was conducted with older adults (aged 75-85). In that experiment, two TBA levels were compared: Low (basic information including only robot-to-human transparency) and High (including additional information relating to predicted outcomes with robot-of-human and human-to-robot transparency). The results revealed a significant difference between the two TBA levels of the model in terms of *perceived usefulness*, *ease of use*, and *attitude*. The High TBA level resulted in improved user acceptance and was preferred by the users.

---

\* Place the footnote text for the author (if applicable) here.

\* Authors contributed equally to this work

# 1 INTRODUCTION

As robots become more and more capable, they are being used in more complicated tasks and environments by both professionals and the general public. In parallel, as robots become more multifaceted, users are becoming less aware of their actions, internal processes and intent [1]. In this context, transparency is an important aspect of human–robot interaction (HRI) [2]. Transparency can be defined as the amount of information the system provides to the user about its current status [3]. With 'transparency' the operators are provided with knowledge of the autonomous agent's behavior, reliability, and intent [2]. Transparency has also been described as a means of sharing intent and awareness between operator and agent [2], which is especially important for non-professional users and bystanders. [4]

The definitions implies that transparency is achieved through the system design and the development of AI systems [5], [6], [7], [8], [9]. The concept of transparency in Artificial Intelligence is closely intertwined with explanations and explainability [10] while adopting an informational perspective (i.e., informing the users about the ongoing processes [11]). This informational perspective, also defined as prospective transparency, informs the users about the data processing and the working of the system upfront. It describes how the AI system reaches decisions in general. Thus, prospective transparency can be seen as an accountability mechanism [8], [12]. Retrospective transparency, another aspect of transparency, refers to post hoc explanations and rationales [8]. It reveals for a specific case how and why a certain decision was reached, or action executed. Retrospective transparency includes the notion of inspectability and explainability. Beyond this, transparency is multifaceted, encompassing interpretability, openness, accessibility, and visibility [7].

Transparency in robot design and deployment involves two dimensions [7]: technical transparency, revealing technical details and functionalities, and social transparency, enabling robots to make their intentions and actions understandable to humans [7]. Transparency has been communicated through different modes, for example, through speech, sound, or images, with human speech being the most natural form of communication [3]. Transparency is necessary to inform the users what the system is doing and discover how and why something went wrong [7] and without it, accountability is impossible [9], [13]. It is important to adapt transparency to the situation specific requirements of the specific context in which the robot is used [14] to build the user's confidence in the robotic system [9], [13], [15]. In a human-robot interaction context, the end user is not the only agent engaging with the system [7]. Since, transparency is not the same for everyone [9], it should consider the expectations and concerns of all involved, the users, developers, and other stakeholders [11]. The general consideration of transparency in HRI enhances the comprehensibility of the robotic system [1], influencing system trust [1], [7], [11], [16] usability [17] and the quality of robot-to-human communication [17]. This leads to improved overall performance and effectiveness [16], [17].

At present, however, most systems still lack transparency [6], [18]. To achieve transparency, the agent should communicate accurate and efficient information to the human operator as briefly as possible and with the right timing, enabling the operator to maintain a proper understanding of the system's operation and its intent and rationale [17].

Currently, the transparency models that have been presented in the HRI literature are based on information provided only by the robot [1], [2], [17], [19]. However, since in robotic systems the user is a significant partner in the interaction [18], it is important to add knowledge related to *human-to-robot* transparency. This study thus focused on the implementation of a new model of transparency which includes both *robot-of-human* and *human-to-robot* transparency (as detailed in section 3) and an examination of its role in HRI and user acceptance (detailed in section 5).

The model was implemented on a robotic system developed for older adults. This target population was chosen for two main reasons: older adults represent a non-expert population [20], and a great deal of effort is being invested in the development of robotic systems to aid older adults in their daily activities [21], [22] to compensate for shortages of care staff for the elderly [23]. In addition, the number of older adults is increasing, and there is a lack of physical training coaches for this population sector [24]. The human personal trainer is thus becoming a scarce resource that is not always available at the needed time and place [25]; for example, during the COVID-19 pandemic the need for social distancing restricted accessibility to coaches. Social robots have the potential to help older adults to meet the need for a personal trainer [26]. In contrast to a human coach, a robotic coach is more flexible in terms of location and time [25]. An additional advantage of such a robot coach is that it can aid in motivating older adults living in their own homes to increase their physical activity [27], [28].

## 2 TRANSPARENCY MODELS: BACKGROUND

The key components of transparency in a human–robot system are knowing the robot's purpose, analyzing the robot's actions, and receiving information from the robot about its analytical processes of decision making and environmental perception [29]. There are a number of different transparency models [2], [30], [31] related to the degree of information provided to the user (i.e., information quantity), and three of these are presented briefly below.

Lynos [30] described a transparency model based on two aspects of information:

- **Robot-to-human transparency:** robot communicates information to the human about its *knowledge of the system and the environment* before, during or after an interaction. For example, if the robot can override the user's actions, the user should know about this and also understand when and why this can occur.
- **Robot-of-human transparency:** robot communicates its awareness about *factors related to the human*. For example, the robot will monitor the user's cognitive workload, so that it can execute an alert or take control when the user is distressed or fatigued.

Another important transparency model is the situation awareness agent-based transparency (SAT) model [31], which guides what information should be communicated to the operator to promote agent transparency. The SAT model can be used to identify the transparency requirements necessary to support the operator's situation awareness and to increase the probability that the operator will rely on the agent. The model embodies three levels, as follows:

- Level 1: information includes the robot's 3ps: purpose, process, and performance.
- Level 2: information is based on the constraint-driven reasoning process of the robot.
- Level 3: information focuses on the projection to future status.

A similar model, presented by Bhaskara et al. [2], describes four different transparency levels, Low, Medium, High and Very high, as follows:

- Low transparency (What) - serves as a baseline and typically contains basic information, a basic plan, and/or the agent's intent.
- Medium transparency (Why) - includes the agent's reasoning, and why it chooses a particular plan/direction.
- High transparency (What next) – supplies additional information that is related to predicted outcomes, predicted consequences, uncertainty, or additional reasoning underlying the agent's decision.
- Very high – supplies information beyond that provided in 'high transparency,' including projected outcomes and uncertainty information in additional to the information.

Transparency in all these models do not take into account the user's performance; communicating only the robot's information without providing feedback may lead the human to view the robot limited in its perception and hence less capable [3], [32], [33].

In contrast to the above models, in this study we aimed to merge transparency information (*robot-to-human*) and robot actions taken based on transparency information received from the human (*robot-of-human* and *human-to-robot*). We investigated whether the action taken by a robot on the basis of transparency, i.e., information received from the human, can improve HRI and user acceptance.

## 3 TRANSPARENCY-BASED ACTION MODEL AND ITS IMPLEMENTATION

We propose a bidirectional transparency model, termed a transparency-based action (TBA) model, which not only provides transparency information (*robot-to-human*) as most current transparency models are based on, but also allows the robot to take actions based on transparency information received from the human (*robot-of-human* and *human-to-robot*, Figure 1).

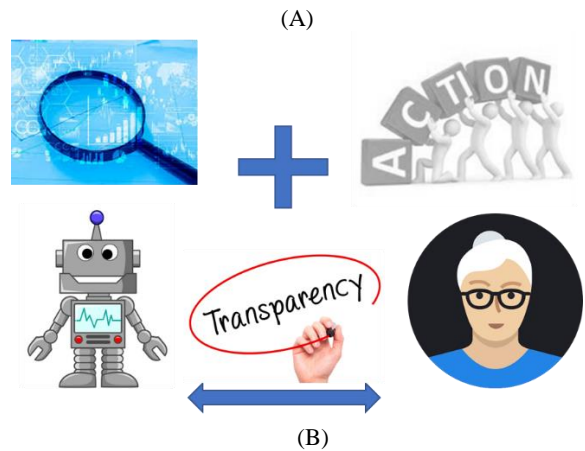

Figure 1: The basics concepts of the TBA model.
(A) Transparency and action (B) Bidirectional transparency between the robot and human.

To create the new transparency-based action (TBA) model (Figure 2) that we implemented in this study - we built on, combined, and complemented the three models described above, as follows: We adopted the terms for three of the four levels of transparency used by Bhaskara et al. [2] (the Very high Transparency level was not used, since our system has no uncertainty information), while using some of the categories of information defined by Chen et al. [31] and some aspects of information transparency presented by Lyons [30]. The TBA model thus involves three levels of robot actions based on bidirectional user↔robot transparency, leading to a more user-centered design of the interaction. The bidirectional transparency was obtained by merging robot-to-human transparency with human-to-robot and robot-of-human transparency, i.e., performance of the human monitor by the robot and information provided by the human to the robot. By providing improved feedback in our model, the system is perceived as more intelligent and hence trustworthy [34].

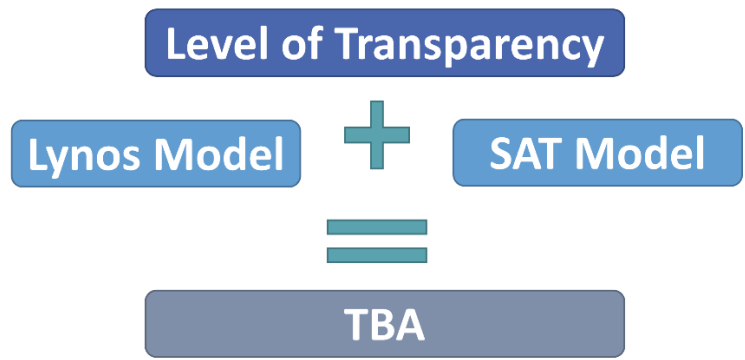

Figure 2: TBA model inspiration from existing transparency models

The model was implemented on a personal robotic trainer for older adults developed in our lab [35]. This robotic trainer for upper body exercise, known as "Gymmy" (Figure 3), was developed to motivate older adults to engage in physical activity. User performance is monitored by skeleton tracking software based on an RGB-D camera, and Gymmy provides feedback during training on the basis of the software's tracking of user performance. Both visual and audio modalities are used to give feedback from the robot to the user, based on previous recommendations [35], [36]. A training session with Gymmy includes several different exercises. Before each exercise, Gymmy provides verbal instructions, including the number of repetitions that should be completed for each exercise. Gymmy demonstrates and performs the exercise as many times as the user requires it to do so. The robot counts each correct repetition that the user completes during this time.

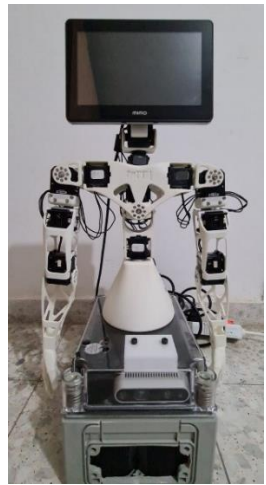

Figure 3: Poppy Torso robot, "Gymmy" used in the study

In the TBA model, each TBA level embodies three aspects of information from the robot (robot actions): purpose (the name of the exercise and an explanation of its purpose), process (an explanation of the procedure, including the number of repetitions required for each exercise), and performance (the progress of the user, i.e., a count of the user's repetitions). Each TBA level contains the information from the previous level, and as the level changes from a lower to a higher level, transparency increases, and the adaption of the physical training program to the user also increases accordingly. The content of the information and action of the robot at each TBA level differs from level to level, as presented below (Figure 4):

- **Low TBA (What)** – basic information predicated only on *robot-to-human* transparency. This level is independent of the robot's knowledge of the human user.

    In Gymmy, this level is implemented by the robot's selection of the number of repetitions at random.

- **Medium TBA (Why)** – the robot's reasoning is added, namely, why it chooses a particular plan/direction for an exercise. This level depends on the information provided by the user to the robot at the beginning of the training, i.e., the model also implements *human-to-robot* transparency. The robot's actions are thus dependent on its knowledge of the human. As a result, the robot chooses its actions based on the information supplied by the human, and provides information on why a specific exercise was chosen.

    In Gymmy, this level is implemented by having the robot define the number of repetitions for each exercise based on the information provided by the user at the beginning of the session.

- **High TBA (Future State)** – additional information is presented, related to predicted outcomes. This level uses previous information provided by the user and the performance of the user in real-time to predict future outcomes for the user, thereby adding *robot-of-human* transparency. The robot's actions are thus dependent on its knowledge of the human.

    In Gymmy, this level is implemented by having the robot define the number of repetitions for each exercise based on the user's performance and on information provided by the user. At the beginning of each session, the robot also explains the entire training session, and gives projections for outcomes.

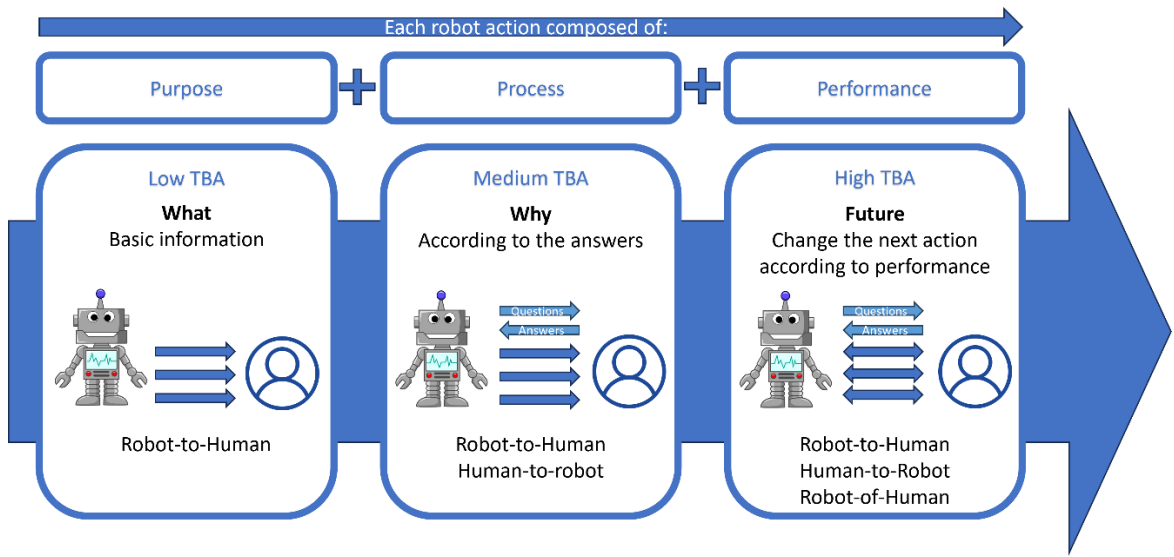

Figure 4: The Low, Medium and High TBA levels

## 4 METHODS

### 4.1 Overview

User studies were conducted to examine the TBA model and its influence on HRI. First, two pilot studies were conducted with engineering students to examine the feasibility and limitations of the proposed TBA model. The pilot studies were followed by an experiment with older adults, the target population of our physical robotic trainer. All studies were approved by our department's ethics committee.

The technology acceptance model (TAM) [37] was used to analyze acceptance of the system, with the independent variable being the TBA level. The dependent variables were subjective measures related to the acceptance of the robotic training.

### 4.2 Hypotheses

Based on previous HRI studies dealing with transparency [1], [2], [17], [30], [38], [39], we propose the following hypotheses (Figure 5).

Older adults will prefer the High TBA level compared to the Low TBA level in terms of:

H1: Perceived usefulness

H2: Ease of use

H3: Attitude

H4: Intention to use

Additionally, we investigated the relations between the original components of the TAM, i.e., perceived usefulness and ease of use with attitude as well as attitude with intention to use [37], as follows:

H5: Higher perceived usefulness of the system will influence a positive attitude to using the system.

H6: Higher ease of use of the system will influence a positive attitude to using the system.

H7: A positive attitude will influence higher behavioral intention to use the system.

Finally, bearing in mind that high performance is the end goal of HRI and that studies have shown that as system transparency increases, human performance also increases [17], [29], we added H8:

H8: older adults will perform better at the High TBA level.

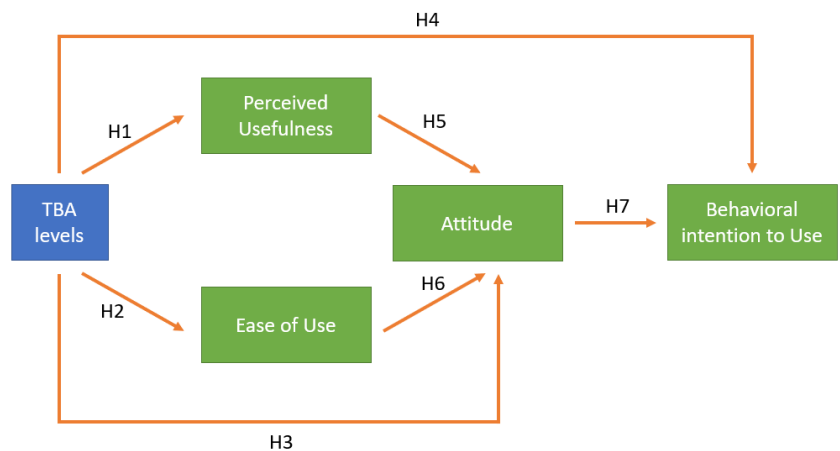

Figure 5: The study hypotheses (based on [37])

### 4.3 Pilot studies

The two pilot studies were conducted to examine the feasibility of the model and to identify its limitations. The students in the pilot studies were recruited via an email sent to all students in an academic course. Students who volunteered to participate in the experiment received a bonus point for that course.

The first pilot study examined the new transparency model with a robotic trainer [40]. The study was conducted with 51 industrial engineering students (16 males, 35 females), aged 23 to 28 (mean = 25.667, std = 1.275). The experiment was performed with a within-subject design, with each volunteer taking part in three sessions with Gymmy, one for each TBA level. The results [41] showed that the students were not able to differentiate significantly between the three TBA levels. Therefore, the medium TBA level was dropped in the subsequent experiments.

In the second pilot study, we examined the model without the Medium TBA [42]. Ten students aged 22-28 years participated in a within-subject study, where each subject performed two sessions with Gymmy, each with a different TBA level. The results [25] revealed that all the participants understood the differences between the two TBA levels. The Wilcoxon signed-rank test revealed that perceived usefulness was the only measure that was significantly different between the levels (P-value = 0.0238, High TBA: mean = 4, std = 0.433 vs. Low TBA: mean = 3.5, std = 0.433). For the two other measures, similar values were obtained for the two levels, as follows. Attitude: Low TBA (mean = 3.731, std = 0.582), High TBA (mean = 3.722, std = 0.516); and ease of use: Low TBA (mean = 4.389, std = 0.486), High TBA (mean = 4.417, std = 0.495).

### 4.4 Main Study

#### 4.4.1 Participants

The experiment was conducted with 21 residents of Palace Lehavim, an older adult independent living facility. The study was publicized via internal social groups at Palace Lehavim two weeks before the planned experiment. The participants were asked to register for the experiment with the local person in charge, either by phone or by using WhatsApp.

#### 4.4.2 Experimental design

The experiment was based on a within-subject design, with each participant performing both Low and High TBA levels. The Low level presented only basic information (including only *robot-to-human* transparency) while the High level included additional information relating to predicted outcomes (with *robot-of-human* and *human-to-robot* transparency). Participants were divided randomly into two groups, and each group performed the experiment in a different order of the TBA levels (half starting with the Low TBA level and the other half starting with the High TBA level).

At the beginning of the experiment, the participants signed an informed consent form and completed a pre-trial questionnaire, which was subsequently ranked using a 7-point Likert scale. The participants were deliberately kept unaware of the details regarding the TBA model during their interactions with the robot. This intentional lack of information was a crucial aspect of our experimental design to minimize the influence and bias.

Each session began with the participant standing in front of the robot. In each session, "Gymmy" demonstrated and explained the exercise it was performing according to the TBA level (Figure 6). Each session comprised the same six exercises presented in random order. In this main study, the timing of the information from the user was set to the middle of the session to make the questions more interactive for the user. The participants completed a final questionnaire at the end of the experiment. A user full session with the Gymmy is shown in Figure 7.

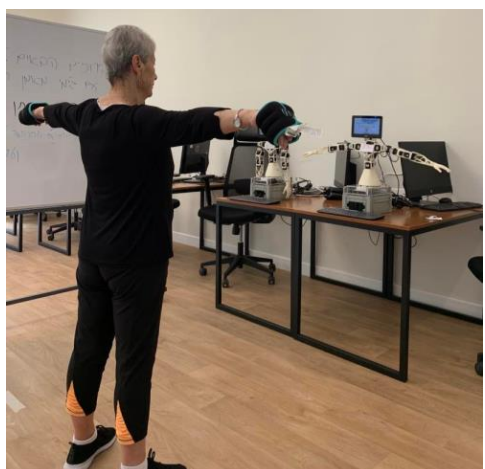

Figure 6: Trainee with "Gymmy"

| | Robot | User |
|---|---|---|
| Introduction | My name is Gymmy1/2 and I will be your personal trainer. We will train together.<br>We're going to do 6 exercises together. You can follow my lead or go at your own pace. I will count every correct repetition.<br>The number of times we'll do each exercise depends on how well you're doing.<br>If you do it right, we'll do more, and if needed, we can go back.<br>I'll also take a break in the middle to check in and learn more about how you're doing. | |
| | When you are ready to start the training, please wave. | The user waves. |
| | Let's go! | |
| Exercise 1 | To target your shoulders, lift both of your arms horizontally. We'll do 6 repetitions. Let's go! | The user made 6, 6 repetitions. |
| | Good job! Because of the success in the previous exercise, I am increasing the number of repetitions in the next exercise. | |
| Exercise 2 | To focus on your biceps, simply bend your elbows. We'll do 8 repetitions. Let's go! | The user made 5,7 repetitions. |
| | Not bad! Even though you didn't pass all the repetitions, I believe in you! We will keep the same number of repetitions in the next exercise. | |
| Exercise 3 | To focus on your shoulders, lift both of your arms at 90 degrees and close forward. We'll do 8 repetitions. Let's go! | The user made 6,8 repetitions. |
| | Good job! Because of the success in the previous exercise, I am increasing the number of repetitions in the next exercise. | |
| Questions | Hope you are enjoying the training, please move closer to the screen to answer some questions. | The user answered C. |
| | Would you like to modify the number of repetitions? The current number of repetitions is 10.<br>A) Increase repetitions    B) Decrease repetitions    C) Keep a fixed number | |
| | What should be the number of repetitions for the upcoming exercises? Options between 4 and 14. | The user chose 10. |
| | Thank you for providing your answers. please go back two steps, and we will continue the training!<br>As requested, the number of repetitions will not change in the following. | |
| Exercise 4 | To target your shoulders, lift both of your arms forward. We'll do 10 repetitions. Let's go! | The user made 5,9 repetitions. |
| | You almost did it! As requested, the number of repetitions will not change in the next exercise but I sure you can do it! | |
| Exercise 5 | To target your shoulders, open both of your arms and bend your elbows. We'll do 10 repetitions. Let's go!" | The user made 6,10 repetitions. |
| | Perfect! As requested, the number of repetitions will not change in the next exercise. | |
| Exercise 6 | To focus on your shoulders, pen arms and bring them forward to the middle. We'll do 10 repetitions. Let's go!" | The user made 6,10 repetitions. |
| | Great work! | |
| Discussion | Thank you for training with me, hope you enjoyed your training. Hope to see you again ! Gymmy1/2. | |
| **Legend**<br>Robot action<br>Low TBA - Gymmy 1<br>High TBA – Gymmy 2 | | |

Figure 7: An example of a robot-user session

### 4.5 Measures

The pre-trial questionnaire was designed to assess the participants' backgrounds (age, gender, education and technology usage) and attitudes to technology and robots [Technology Assessment Property (TAP) and Negative Attitude towards Robots (NARS) questionnaires]. The TAP questionnaire [43] examines the participant's level of technological knowledge and includes questions grouped into three sections: optimism regarding technology, confidence in using new technology, and reliance on technology (Appendix A). The NARS questionnaire [44] examines the participants' perception of and attitude to robots. It comprises three sections related to negative attitude toward: situations of interaction with robots (S1), the social impact of robots (S2), and emotions in interactions with robots (S3). In the current study we employed only 2-3 questions from each section so as not to overload the older adults; the specific questions were selected corresponding to previous studies [35], [36] (Appendix B).

The final questionnaire and measures are presented in Table 1. The following subjective measures based on the TAM were assessed on a Likert 7-point scale: perceived usefulness, ease of use (comfortability and understanding), attitude (engagement, trust, satisfaction and enjoyment), and intention to use. The objective measure was the success rate in the physical exercises. Success was defined for each exercise during the training as the number of successful repetitions divided by the total number of repetitions in the exercise. The total success rate of the training was calculated as the average value.

Table 1: Post trial questionnaire and measures

| Dependent variable | | Question |
|---|---|---|
| Perceived usefulness | | I would be willing to train with the robot again because the training was of value to me |
| | | I felt the training was useful to me |
| Ease of use | Comfortability | I felt comfortable during the interaction |
| | | I felt nervous during the activity |
| | Understanding | I feel the robot understands me |
| | | I understood the information the robot presented to me |
| Attitude | Engagement | I concentrated on the activity for the entire session |
| | | I find the robot pleasant to interact with |
| | Trust | I felt I could really trust this robot |
| | | I can trust the information provided by the robot |
| | | I would trust the robot if it were to give me advice |
| | Satisfaction | I was satisfied by the robot's performance during this activity |
| | | Interacting with the robot was a pleasant and satisfactory experience |
| | | The movements of a robot are unsatisfactory |
| | Enjoyment | I enjoy exercising with the robot |
| | | I was eager to finish the exercise |
| Intention to use | | I would like to exercise with the robot in the future. |
| | | I think that I would like to use this system often |
| Success rate | | Success rate per exercise $i = \frac{\text{number of repetitions of the user}}{\text{Total repetitions for } i}$ |
| | | Success rate $= \frac{\text{sum of success rate per exercise}}{\text{Sum of exercises}}$ |

### 4.6 Analysis

The raw data was pre-processed using Python programming. The analysis of the influence of the dependent variables was evaluated with the Wilcoxon signed-rank test using Python programing. Spearman's rank correlation to evaluate the relationship between the TAM variables was determined with R programming language with RStudio software.

## 5 RESULTS

### 5.1 Participants

Twenty-one older adults aged between 75 and 85 years participated in the experiment (mean age = 80.33, std = 2.82, Figure 8); 9 were women (mean age = 79.677, std = 3) and 12, men (mean age = 80.83, std = 2.7).

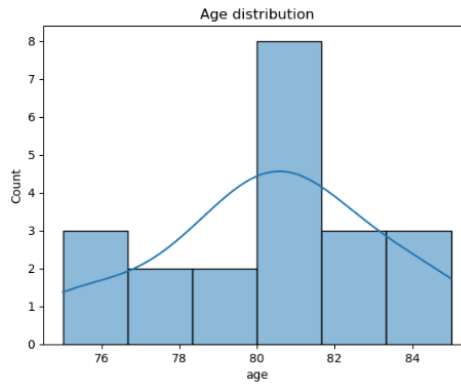

Figure 8: Age distribution of the participants

Of the participants, 23.8% had a high school education and 33.4% a university degree, and the rest had acquired additional education in courses or as practical engineers (Figure 9). Among the participants with degrees, only one held a degree in a field outside of STEM subjects.

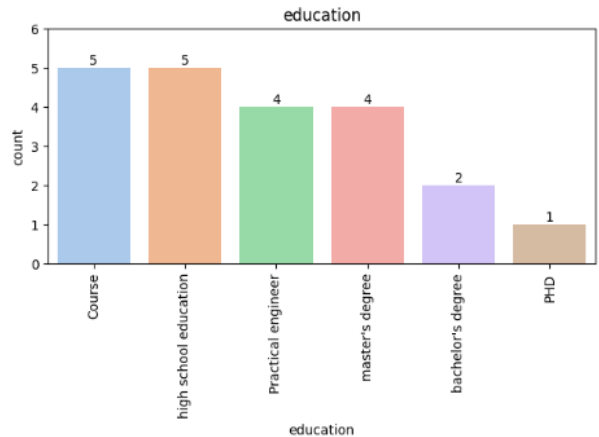

Figure 9: Education distribution of the participants

The majority of the participants were pensioners, but 14.2% were still working. Most of the participants (86%) were physically active, with 67% engaging in sporting activities every day, and 19%, every 2 or 3 days; only 5% almost never engaged in sport (Figure 10).

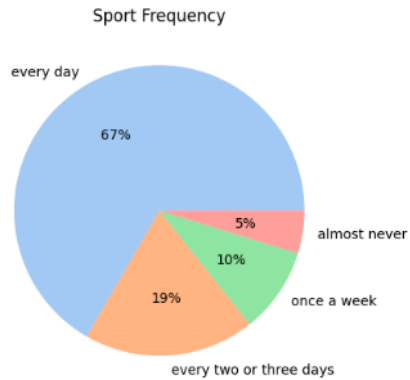

Figure 10: Sport frequency of the participants

In terms of technology use, all the participants used smartphones, 80% also used computers, and 15% also used tablets. About a third (28%) of the participants had interacted with robots in a previous experiment.

### 5.2 TAP

The TAP results revealed that 66.67% pf the participants expressed high **optimism about technology**, while the other 33.33% expressed medium optimism. There were 47.6% of participants with high **confidence in using new technology**, 38.1% with medium confidence, and 14.3% with low confidence. Regarding **dependence on technology**, 14.2% of participants were highly dependent, 55.4% were moderately dependent, and 33.4% were barely dependent on technology.

### 5.3 NARS

The NARS results revealed that 76.1% of the participants had negative feelings about situations and interactions with robots (S1) at a low level, and the rest at a medium level. The results also revealed that 42.85% had a medium level negative attitude toward the social impact of robots (S2), with the remaining participants being divided equally between low and high levels. Finally, 47.6% of the participants had a low level negative attitude toward emotions in interactions with robots (S3), and 42.85% ranked the measure as medium level.

### 5.4 Technology acceptance model

The TBA comparisons were conducted only for participants who distinguished between the two TBA levels, namely, for the 81% of the participants who understood the difference between the levels. Four participants who did not understand the difference were not included in the analysis, and an additional two participants were removed because of technical issues. For each measure, the Wilcoxon signed-rank test was used to determine whether there were differences between the two levels (Table 2).

Table 2: Results for the TAM questionnaire administered at the end of the experiment

| Dependent variable | | Low TBA | | High TBA | | P-value | |
|---|---|---|---|---|---|---|---|
| Perceived usefulness | | 5.033 ± 1.31 | | 5.767 ± 1.389 | | 0.011* | |
| Ease of use | Comfortability | 6.767 ± 0.442 | 6.017 ± 0.793 | 6.8 ± 0.44 | 6.2 ± 0.6904 | 0.317 | 0.011* |
| | Understanding | 5.267 ± 1.377 | | 5.6 ± 1.186 | | 0.011* | |
| Attitude | Engagement | 6.533 ± 0.591 | 6.04 ± 0.86 | 6.633 ± 0.531 | 6.2267 ± 0.7523 | 0.18 | 0.027* |
| | Trust | 5.622 ± 1.382 | | 5.777 ± 1.153 | | 0.109 | |
| | Satisfaction | 6.133 ± 0.925 | | 6.2 ± 0.917 | | 0.18 | |
| | Enjoyment | 6.033 ± 1.024 | | 6.533 ± 0.694 | | 0.039* | |
| Intention to use | | 5.8 ± 1.249 | | 6.133 ± 1.284 | | 0.068 | |

*5.4.1 Perceived usefulness*

According to the questionnaire, most participants found the robot useful (mean = 5.4, std = 1.3988). The Wilcoxon signed-rank test revealed that *Perceived usefulness* was significantly different between the Low TBA and High TBA levels (P-value = 0.0106) (Figure 11). In addition, Spearman's rank correlation revealed a positive correlation between *Perceived usefulness* and *Attitude* ($\rho = 0.636$, P-value = 0.0002).

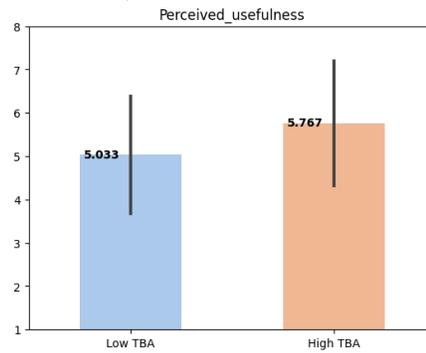

Figure 11: Perceived usefulness mean value for each TBA level

*5.4.2 Ease of use*

According to the questionnaire, the participants found the robot very easy to use (mean = 6.1, std = 0.745). The Wilcoxon signed-rank test revealed that *Ease of use* was significantly different between the Low TBA and High TBA levels (P-value = 0.0114) (Figure 12). In addition, Spearman's rank correlation revealed a positive correlation between *Ease of use* and *Attitude* ($\rho = 0.645$, P-value = 0.00016).

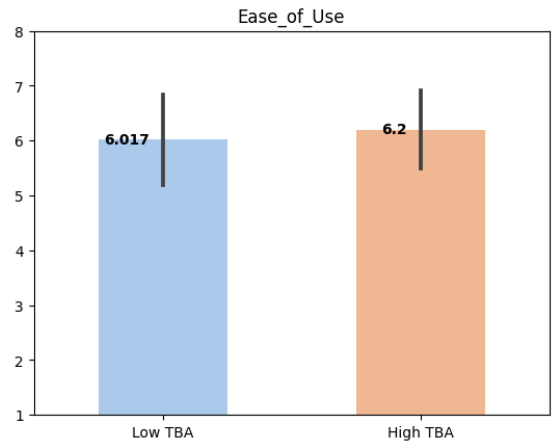

Figure 12: Ease of use mean value for each TBA level

*5.4.3 Attitude*

The participants had a positive attitude towards the robot, with a mean of 6.133 (std = 0.813). The Wilcoxon signed-rank test revealed that *Attitude* was significantly different between the Low TBA and High TBA levels (P-value = 0.0273) (Figure 13). Spearman's rank correlation revealed a positive correlation between the *Attitude* and *Intention to use* measures ($\rho$ = 0.378, P-value = 0.04).

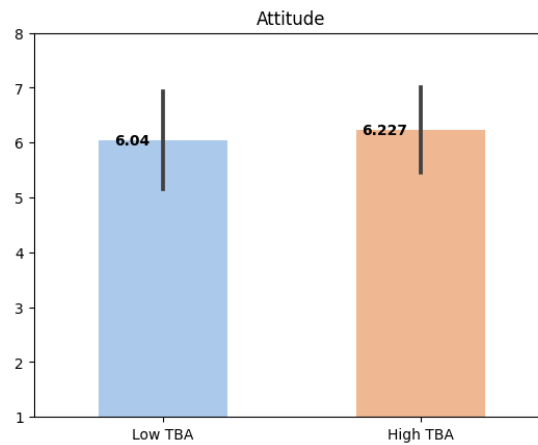

Figure 13: Attitude mean value for each TBA level

*5.4.4 Intention to use*

Most participants indicated that they wanted to use the robot again. The *Intention to use* was very high for both levels (mean = 5.96, std = 1.27); the participants liked the robot and noted that they wanted to use it, independently of the TBA level. The Wilcoxon signed-rank test did not reveal a significant statistical difference between the Low TBA and High TBA levels (P-value = 0.0679) (Figure 14).

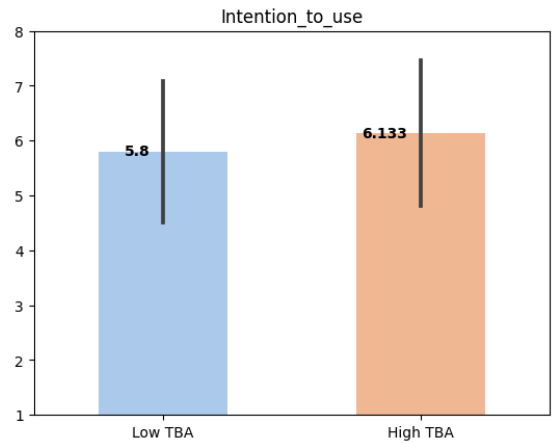

Figure 14: Intention to use mean value for each TBA level

### 5.5 Success Rate

Analysis of the participants' *success rates* revealed that the High TBA improved their success, although results were not statistically different for the two levels (P-value = 0.068). The mean of the *success rate* of the TBA levels was higher for the High TBA (Figure 15a) and also in most cases (80%) for each individual participant (Figure 15b).

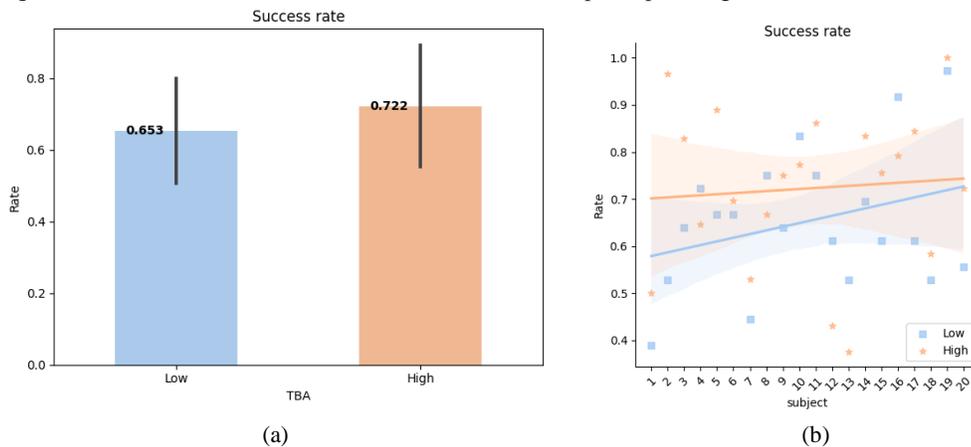

Figure 15 – (a) Mean Success rate, (b) Individual success rate

### 6 DISCUSSION

The evaluation of our TBA model with an older adult population yielded noteworthy results. The findings revealed that a High TBA improved the acceptance of the robot physical trainer in terms of *Perceived usefulness* (P-value = 0.0106), *Ease of use* (P-value = 0.0114) and *Attitude* (P-value = 0.0273), validating H1, H2, and H3, respectively. The results also showed that the majority of the older adults liked the robot and wanted to use it independently of the TBA level (refuting H4 that older adults prefer to use the robot at a High TBA level). Spearman's rank correlation was used to analyze the TAM relationship. All three parts of the relationship (*perceived usefulness* vs. *attitude, ease of use* vs. *attitude* and *attitude* vs. *intention to use*) were positively correlated (validating H5, H6 and H7, respectively). In our system, exercise success is signaled to the user by the system's feedback. When the user performs the movement correctly, the robot counts the correct repetition with audio feedback. The *success rate* in the physical exercises improved for the High TBA (Low TBA mean = 0.653, High TBA mean = 0.733), although the difference was not significant (partially validating H8).

In the developed TBA model, each TBA level embodies three aspects of information from the robot related to robot actions: purpose, process and performance. Each TBA level contains information from the previous level, and as the level changes from a lower to a higher level, transparency increases. The content of the information and action of the robot at each TBA level differs from level to level. We proposed a model with three levels. However, the early pilot studies revealed certain drawbacks in the proposed three level model. The results with engineering students as users revealed they were unable to distinguish significantly between the three TBA levels. This raised concerns about the model's effectiveness. In response to these findings, we conducted a second pilot study, in which we eliminated the Medium TBA level to simplify the model. This adjustment allowed us to observe improved participant acceptance and differentiation between the remaining two TBA levels. These preliminary results played an instrumental role in shaping our research trajectory.

It was important to consider the potential challenges of multiple TBA implementations when preparing for experiments with older adults. To begin with, recruiting enough older adults for an experiment featuring multiple TBA levels proved challenging. Secondly, from a statistical perspective, we recognized that employing a simplified model with fewer TBA levels could yield more robust statistical outcomes. This, in turn, had the potential to enhance the statistical power of our study, augmenting our ability to detect and reveal significant differences. While we entertained the possibility that older adults might successfully differentiate between all three TBA levels, we decided to keep with the two-level model configuration (without the Medium TBA level). This modification enabled a more straightforward and manageable study with older adults. Additionally, this move was driven by the awareness that conducting experiments, particularly with specialized populations such as older adults, is a resource-intensive undertaking, necessitating substantial investments in terms of time, budget, and personnel. By reducing the number of TBA levels, we streamlined the data collection process, making it more feasible within the available resources. This corresponds also to previous research [33] which revealed that older adults are not able to distinguish between more than two levels. This might be due to the novelty effect and might change after getting acquainted with the system. Furthermore, the decision to remove the Medium level can also be framed within the context of an iterative approach to model development. By commencing with a simpler model, we were afforded the opportunity to gather evidence and insights from our experiments with older adults, which, in turn, could inform future iterations of the TBA model. Subsequently, our experiment with the older adult population exhibited a clearer and more distinct contrast between the two remaining TBA levels, ultimately leading to significant findings as discussed above.

## 7 LIMITATIONS AND FUTURE WORK

The present study was limited to one specific social robot, and thus its findings cannot be generalized to other robotic systems. The full model should be evaluated with different social robots and on different tasks. The scope could extend to encompass a diverse range of robotic systems, each custom-made to specific tasks involving varying degrees of risk and/or uncertainty or operating under a multitude of stress conditions (e.g., time, accuracy). One example of an ongoing study is implementing the TBA model in navigation of a tele-operated mobile robot. The robot was equipped with two control modes that enable to switch between levels of automation (Robot-Initiated and Humam-Initiated). The Low TBA is implemented so that the robot chooses the control mode randomly and the High TBA level is implemented so the robot chooses the control mode according to the information provided by the user and according to the user's performance.

Furthermore, only a specific population was evaluated. This population is, however, particularly important, since it represents nonprofessional (non-expert) users. Expanding evaluation of the model with additional populations should be pursued. User studies should be conducted with other types of non-professionals (e.g., individuals with differing technological affinities)—this will enable to distinguish the model's capabilities to address the unique needs of each group. It might be that the TBA level should be personalized based on users' characteristics (e.g., experience, background) and fitted to specific conditions (task, robot, environment). Furthermore, different stakeholders (e.g., developer, user, deployer) have different roles and information needs, background knowledge and abilities. Accordingly, the transparency requirement needs to be tailored both to the types of users (in light of their roles, responsibilities, and interests) and to their level of capacity and vulnerability [7].

A longitudinal study is needed to capture changes in acceptance as well as the behavior that develops over time, which can influence assimilation. The effectiveness of the TBA model along time, after it has been used in real-life situations for a while and overcoming the novelty affect, might influence the impact of the transparency. Furthermore, the need for different TBA levels might evolve over time (influenced by factors such as user experience, fatigue or boredom, e.g., complex vs simple, certainty levels, failures or environmental changes e.g., new/unknown vs familiar, complexity, dynamic vs static environment). This might require dynamic change between TBA levels leading to the need to develop an adaptive model that is able to switch between TBA levels depending on user, robot, task and environment conditions.

This work focused on the evaluation of the TBA model using the 'Technology Acceptance Model'. Future research should consider incorporating a broader range of evaluation metrics to provide a more comprehensive understanding of the model's impact on various facets of user experience and performance.

In addition to exploring the role of transparency in making robotic systems more understandable [5], future work should also explore another dimension that can enhance understandability - explainability [45]. Explainability refers to explicitly providing the rationale for the system's behavior depending on the user's need and his/her knowledge [46], also known as retrospective transparency as mentioned in the background. Prospective transparency is defined as the information provided to the users about the data processing and the working of the system upfront with projection to a future state corresponding to our proposed High TBA level. Research should aim to evaluate how understandability is affected by both transparency and explainability (e.g. prospective and retrospective transparency) and how each of these dimensions impact user acceptance and HRI. Ideally, this should be implemented with additional levels of transparency, linking understanding, and explainability (similar to the 10 levels of automation [47]). However, this requires investigating into the number of levels applied in a specific application and perhaps ways to select the best set of levels since in a specific application it is recommended to utilize 2-3 levels only.

A crucial aspect not explored within our research is the legal and ethical dimensions of transparency. It is imperative to underscore the significance of integrating legal frameworks and ethical considerations into the development and utilization of robotic systems [7]. Addressing these diverse dimensions of transparency is essential for cultivating trust and promoting the acceptance of robotic technologies in society [7], [11]. Future work should delve into these legal and ethical aspects to provide a comprehensive understanding of transparency in robotic systems.

## 8 CONCLUSIONS

Existing transparency models are based only on information provided by the robot and do not take the user into consideration. We developed a **bidirectional transparency model** for a social assistive robot that merges transparency information from the robot (*robot-to-human*) with robot actions taken on the basis of transparency information received from the human (*robot-of-human* and *human-to-robot*). The results of this study with older adults using a robotic physical trainer revealed that the TBA model can lead to **increased acceptance** by the user. The Wilcoxon signed-rank test revealed that *Perceived usefulness*, *Ease of use*, and *Attitude* were significantly different between the two TBA levels of the model, with users preferring the High TBA. The ability to provide bidirectional transparent information establishes an important milestone toward acceptance of robotic systems working in collaboration with humans.

This unique bidirectional transparency makes the TBA model stand out from existing transparency models fostering a more comprehensive and interactive exchange between the user and the robot. The in-depth examination of transparency levels within the TBA model and their discernible impact on user experience yields valuable insights for the design and implementation of human-robot interaction systems. By emphasizing the relationship of transparency elements and their influence on user acceptance, our study results contribute meaningful perspectives that can inform the development of future human-robot interaction frameworks. Also by addressing the user and providing transparency based on the user's actions, we provide a step forward in development of an adaptive system. This bidirectional transparency enhances the robot's perception leading to a more intelligent system.

## APPENDICES

### APPENDIX A. TAP questions

| | | TAP - Technology Assessment Property |
|---|---|---|
| Optimism in technology | 1 | Technology gives me more control over my daily life. |
| | 2 | Technology helps me make more necessary changes in my life. |
| | 3 | New technologies make my life easier. |
| The confidence of the person to use new technology | 4 | I can figure out new high-tech products and services without help from others. |
| | 5 | I seem to have fewer problems that other in making technology work. |
| | 6 | I enjoy figuring out how to use new technologies. |
| Reliance on technology | 7 | Technology controls my life more than I control technology. |

### APPENDIX B. NARS questions

| | | NARS - negative attitude towards robots |
|---|---|---|
| S1: Negative attitude toward situations of interaction with robots. | 1 | I would feel uneasy if I was given a job where I had to use robots. |
| | 2 | I would feel nervous operating a robot in front of other people. |
| | 3 | I would feel very nervous just standing in front of robot |
| S2: Negative attitude toward social influence of robots | 4 | I would feel uneasy if robots really had emotions. |
| | 5 | I feel that if I depend on robots too much, something bad might happen. |
| S3: Negative attitude toward emotions in interaction with robots | 6 | I would feel relaxed talking with robots (*) |
| | 7 | if robots had emotions, I would be able to make friends with them (*). |

(*) - Positive sentences were reversed when calculating the measure.

# REFERENCES

[1] B. Nesset, D. A. Robb, J. Lopes, and H. Hastie, "Transparency in HRI: Trust and decision making in the face of robot errors," *ACM/IEEE International Conference on Human-Robot Interaction*, pp. 313–317, 2021, doi: 10.1145/3434074.3447183.

[2] A. Bhaskara, M. Skinner, and S. Loft, "Agent Transparency: A Review of Current Theory and Evidence," *IEEE Trans Hum Mach Syst*, vol. 50, no. 3, pp. 215–224, 2020, doi: 10.1109/THMS.2020.2965529.

[3] Tracy L. Sanders T, Tarita Wixon, K. Elizabeth Schafer, Jessie Y.C Chen, and P.A Hancock, "The Influence of Modality and Transparency on Trust in Human-Robot Interaction," IEEE International Inter-Disciplinary Conference on Cognitive Methods in Situation Awareness and Decision Support (CogSIMA), San Antonio, TX, USA, 2014, pp. 156–159. doi: 10.1109/CogSIMA.2014.6816556.

[4] J. Tullio, A. K. Dey, J. Chalecki, and J. Fogarty, "How it works: A field study of non-technical users interacting with an intelligent system," *Conference on Human Factors in Computing Systems - Proceedings*, pp. 31–40, 2007, doi: 10.1145/1240624.1240630.

[5] A. Barredo Arrieta *et al.*, "Explainable Artificial Intelligence (XAI): Concepts, taxonomies, opportunities and challenges toward responsible AI," *Information Fusion*, vol. 58, no. October 2019, pp. 82–115, 2020, doi: 10.1016/j.inffus.2019.12.012.

[6] M. Eiband, H. Schneider, M. Bilandzic, J. Fazekas-Con, M. Haug, and H. Hussmann, "Bringing transparency design into practice," *International Conference on Intelligent User Interfaces, Proceedings IUI*, pp. 211–223, 2018, doi: 10.1145/3172944.3172961.

[7] H. Felzmann, E. Fosch-Villaronga, C. Lutz, and A. Tamo-Larrieux, "Robots and Transparency: The Multiple Dimensions of Transparency in the Context of Robot Technologies," *IEEE Robotics and Automation Magazine*, vol. 26, no. 2. Institute of Electrical and Electronics Engineers Inc., pp. 71–78, Jun. 01, 2019. doi: 10.1109/MRA.2019.2904644.

[8] H. Felzmann, E. Fosch-Villaronga, C. Lutz, and A. Tamò-Larrieux, "Towards Transparency by Design for Artificial Intelligence," *Sci Eng Ethics*, vol. 26, no. 6, pp. 3333–3361, Dec. 2020, doi: 10.1007/s11948-020-00276-4.

[9] A. F. T. Winfield *et al.*, "IEEE P7001: A Proposed Standard on Transparency," *Front Robot AI*, vol. 8, Jul. 2021, doi: 10.3389/frobt.2021.665729.

[10] M. Veale, M. Van Kleek, and R. Binns, "Fairness and accountability design needs for algorithmic support in high-stakes public sector decision-making," in *Conference on Human Factors in Computing Systems - Proceedings*, Association for Computing Machinery, Apr. 2018. doi: 10.1145/3173574.3174014.

[11] H. Felzmann, E. F. Villaronga, C. Lutz, and A. Tamò-Larrieux, "Transparency you can trust: Transparency requirements for artificial intelligence between legal norms and contextual concerns," *Big Data Soc*, vol. 6, no. 1, Jan. 2019, doi: 10.1177/2053951719860542.

[12] J. Zerilli, A. Knott, J. Maclaurin, and C. Gavaghan, "Transparency in Algorithmic and Human Decision-Making: Is There a Double Standard?," *Philos Technol*, vol. 32, no. 4, pp. 661–683, Dec. 2019, doi: 10.1007/s13347-018-0330-6.

[13] A. Weller, "Transparency: Motivations and Challenges," *Explainable AI: interpreting, explaining and visualizing deep learning. Cham: Springer International Publishing, 2019. 23-40.*.

[14] K. Fischer, H. M. Weigelin, and L. Bodenhagen, "Increasing trust in human-robot medical interactions: Effects of transparency and adaptability," *Paladyn J Behav Robot 9 (1)*, vol. 9, no. 1, pp. 95–109, Feb. 2018, doi: 10.1515/pjbr-2018-0007.

[15] M. M. A. D. E. Graaf, A. Dragan, B. F. Malle, and T. O. M. Ziemke, "Introduction to the Special Issue on Explainable," *ACM Trans Hum Robot Interact*, vol. 10, no. 3, 2021, doi: 10.1145/3461597.

[16] P. A. Hancock, D. R. Billings, K. E. Schaefer, J. Y. C. Chen, E. J. De Visser, and R. Parasuraman, "A meta-analysis of factors affecting trust in human-robot interaction," *Hum Factors*, vol. 53, no. 5, pp. 517–527, Oct. 2011, doi: 10.1177/0018720811417254.

[17] J. E. Mercado, M. A. Rupp, J. Y. C. Chen, M. J. Barnes, D. Barber, and K. Procci, "Intelligent Agent Transparency in Human-Agent Teaming for Multi-UxV Management," *Hum Factors*, vol. 58, no. 3, pp. 401–415, 2016, doi: 10.1177/0018720815621206.

[18] M. R. Endsley, "From Here to Autonomy: Lessons Learned from Human-Automation Research," *Hum Factors*, vol. 59, no. 1, pp. 5–27, 2017, doi: 10.1177/0018720816681350.

[19] J. Y. C. Chen and M. J. Barnes, "Agent Transparency for Human-Agent Teaming Effectiveness," pp. 1381–1385, 2015, doi: 10.1109/SMC.2015.245.

[20] S. J. Czaja and J. Sharit, "Age differences in attitudes toward computers," *Journals of Gerontology - Series B Psychological Sciences and Social Sciences*, vol. 53, no. 5, pp. 329–340, 1998, doi: 10.1093/geronb/53B.5.P329.

[21] H. Lee, C. Yang, F. Lin, and P. Yang, "Age Difference in Perceived Ease of Use , Curiosity ," *ACM Trans Hum Robot Interact*, vol. 8, no. 2, 2019, doi: 10.1145/3311788.

[22] S. J. Czaja, W. R. Boot, N. Charness, and W. A. Rogers, *Designing for Older adults*. Principles and Creative Human Factors Approaches, Third Edition (3rd ed.). CRC Press, 2019. doi: https://doi.org/10.1201/b22189.

[23] P. I. Buerhaus, "Current and Future State of the US Nursing Workforce," *JAMA*, vol. 300, no. 20, p. 2422, Nov. 2008, doi: 10.1001/jama.2008.729.

[24] A. Lotfi, C. Langensiepen, and S. Yahaya, "Socially Assistive Robotics: Robot Exercise Trainer for Older Adults," *Technologies (Basel)*, vol. 6, no. 1, p. 32, 2018, doi: 10.3390/technologies6010032.

[25] B. Görer, A. A. Salah, and H. L. Akın, "An autonomous robotic exercise tutor for elderly people," *Auton Robots*, vol. 41, no. 3, pp. 657–678, Mar. 2017, doi: 10.1007/s10514-016-9598-5.

[26] J. Fasola and M. Mataric, "A Socially Assistive Robot Exercise Coach for the Elderly," *J Hum Robot Interact*, vol. 2, no. 2, pp. 3–32, 2013, doi: 10.5898/jhri.2.2.fasola.

[27] E. Ruf, S. Lehmann, and S. Misoch, "Motivating older adults to exercise at home: Suitability of a humanoid robot," *ICT4AWE 2020 - Proceedings of the 6th International Conference on Information and Communication Technologies for Ageing Well and e-Health*, no. Ict4awe, pp. 107–114, 2020, doi: 10.5220/0009341001130120.

[28] M. Čaić, J. Avelino, D. Mahr, G. Odekerken-Schröder, and A. Bernardino, "Robotic Versus Human Coaches for Active Aging: An Automated Social Presence Perspective," *Int J Soc Robot*, vol. 12, no. 4, pp. 867–882, 2020, doi: 10.1007/s12369-018-0507-2.

[29] S. Guznov *et al.*, "Robot Transparency and Team Orientation Effects on Human–Robot Teaming," *Int J Hum Comput Interact*, vol. 36, no. 7, pp. 650–660, 2020, doi: 10.1080/10447318.2019.1676519.

[30] J. B. Lyons, "Being transparent about transparency: A model for human-robot interaction," *AAAI Spring Symposium - Technical Report*, vol. SS-13-07, pp. 48–53, 2013.

[31] J. Y. C. Chen, K. Procci, M. Boyce, J. Wright, A. Garcia, and M. J. Barnes, "Situation Awareness–Based Agent Transparency," *US Army Research Laboratory*, no. April, pp. 1–29, 2014.

[32] S. Schneider and F. Kummert, "Comparing Robot and Human guided Personalization: Adaptive Exercise Robots are Perceived as more Competent and Trustworthy," *Int J Soc Robot*, vol. 13, no. 2, pp. 169–185, 2021, doi: 10.1007/s12369-020-00629-w.

[33] S. Olatunji, T. Oron-Gilad, V. Sarne-Fleischmann, and Y. Edan, "User-centered feedback design in person-following robots for older adults," *Paladyn*, vol. 11, no. 1, pp. 86–103, Jan. 2020, doi: 10.1515/pjbr-2020-0007.

[34] S. Krening and K. M. Feigh, "Characteristics that influence perceived intelligence in AI design," in *Proceedings of the Human Factors and Ergonomics Society*, Human Factors and Ergonomics Society Inc., 2018, pp. 1637–1641. doi: 10.1177/1541931218621371.

[35] M. Krakovski *et al.*, "'Gymmy': Designing and testing a robot for physical and cognitive training of older adults," *Applied Sciences (Switzerland)*, vol. 11, no. 14, 2021, doi: 10.3390/app11146431.

[36] O. Avioz-Sarig, S. Olatunji, V. Sarne-Fleischmann, and Y. Edan, "Robotic System for Physical Training of Older Adults," *Int J Soc Robot*, vol. 13, no. 5, pp. 1109–1124, 2021, doi: 10.1007/s12369-020-00697-y.

[37] F. D. Davis, "Perceived usefulness, perceived ease of use, and user acceptance of information technology," *MIS Q*, vol. 13, no. 3, pp. 319–339, 1989, doi: 10.2307/249008.

[38] T. Reid, "Improving Human-Machine Collaboration Through Transparency-based Feedback – Part I: Human Trust and Workload Model," *IFAC-PapersOnLine*, vol. 51, no. 34, pp. 315–321, 2019, doi: 10.1016/j.ifacol.2019.01.028.

[39] S. Ososky *et al.*, "Determinants of system transparency and its influence on trust in and reliance on unmanned robotic systems," no. June 2014, 2014, doi: 10.1117/12.2050622.

[40] M. Krakovski, N. Aharony, and Y. Edan, "Robotic Exercise Trainer : How Failures and T-HRI Levels Affect User Acceptance and Trust," in SCRITA Workshop Proceedings (arXiv:2208.11090) held in conjunction with 31st IEEE International Conference on Robot & Human Interactive Communication, 29/08 - 02/09, Naples (Italy), 2022, pp. 1–4.

[41] N. Aharony, "TBA model," Ben-Gurion University of the Negev, Beer Sheva, 84105, Israel, 2023.

[42] N. Aharony, M. Krakovski, and Y. Edan, "Linking Actions to Transparency for Improved HRI – Pilot Study," *International Conference on Social Robotics (ICSR)*, pp. 4–6, 2022.

[43] M. Ratchford and M. Barnhart, "Development and validation of the technology adoption propensity (TAP) index," *J Bus Res*, vol. 65, no. 8, pp. 1209–1215, 2012, doi: 10.1016/j.jbusres.2011.07.001.

[44] D. S. Syrdal, K. Dautenhahn, K. L. Koay, and M. L. Walters, "The Negative Attitudes towards Robots Scale and Reactions to Robot Behaviour in a Live Human-Robot Interaction Study," *Adaptive and emergent behaviour and complex system*, 2009.

[45] S. Wallkötter, S. Tulli, G. Castellano, A. Paiva, and M. Chetouani, "Explainable Embodied Agents through Social Cues: A Review," *ACM Trans Hum Robot Interact*, vol. 10, no. 3, 2021, doi: 10.1145/3457188.

[46] Hoffman. Robert R, Mueller. Shane T, K. Gary, and L. Jordan, "Metrics for Explainable AI: Challenges and Prospects," pp. 1–50, 2018.

[47] M. R. Endsley and D. B. Kaber, "Level of automation effects on performance, situation awareness and workload in a dynamic control task," *Ergonomics*, vol. 42, no. 3, pp. 462–492, Mar. 1999, doi: 10.1080/001401399185595.